\documentclass[runningheads]{llncs}
\usepackage[T1]{fontenc}
\usepackage{multirow}

\usepackage{graphicx}
\usepackage{booktabs}
\usepackage{subfigure}
\usepackage[accsupp]{axessibility}

\begin{document}
\title{ AGCD-Net: Attention Guided Context Debiasing Network for Emotion Recognition}
\titlerunning{AGCD-Net for Emotion Recognition}
%
\author{Varsha Devi\inst{1}\orcidID{0000-0003-4581-0149} \and
Amine Bohi\inst{2}\orcidID{0000-0002-2435-3017} \and
Pardeep Kumar\inst{3}\orcidID{0000-0003-1590-665X}}
\authorrunning{Devi et al.}
%
\institute{CESI LINEACT Laboratory, UR 7527, Nice, 06200, France, \email{vdevi@cesi.fr}  \and
CESI LINEACT Laboratory, UR 7527, Dijon, 21800, France, \email{abohi@cesi.fr}\\
\and
ACENTAURI Team, Inria Centre at Universit\'e C\^ote d’Azur, Sophia Antipolis, 06560, France, \email{pardeep.kumar@inria.fr}}
\maketitle              
\begin{abstract}
Context-aware emotion recognition (CAER) enhances affective computing in real-world scenarios, but traditional methods often suffer from context bias-spurious correlation between background context and emotion labels (e.g. associating ``garden'' with ``happy''). In this paper, we propose \textbf{AGCD-Net}, an Attention Guided Context Debiasing model that introduces \textit{Hybrid ConvNeXt}, a novel convolutional encoder that extends the ConvNeXt backbone by integrating Spatial Transformer Network and Squeeze-and-Excitation layers for enhanced feature recalibration. At the core of AGCD-Net is the Attention Guided - Causal Intervention Module (AG-CIM),  which applies causal theory, perturbs context features, isolates spurious correlations, and performs an attention-driven correction guided by face features to mitigate context bias. Experimental results on the CAER-S dataset demonstrate the effectiveness of AGCD-Net, achieving state-of-the-art performance and highlighting the importance of causal debiasing for robust emotion recognition in complex settings.
\end{abstract}

\section{Introduction}
\label{sec:intro}

Emotion recognition is a key task in artificial intelligence, with applications including but not limited to health~\cite{drimalla2020towards,bohi2024novel}, human-robot interaction~\cite{icaart25},
and social robotics~\cite{greco2019emotion}. It involves classifying input into categories such as \emph{angry}, \emph{sad}, \emph{happy}, etc. Traditional methods rely on single modalities (e.g., facial analysis~\cite{el2023emonext}, body postures~\cite{ahmed2019emotion}, or acoustic cues~\cite{schuller2009acoustic}), or on multimodal fusion~\cite{yan2024multimodal}, but they often fail in unconstrained environments due to pose variations, occlusions, or overreliance on facial cues ~\cite{bohi2024novel,icaart25,gursesli2024facial}.

Context-Aware Emotion Recognition (CAER) has emerged to address these issues by jointly leveraging facial and contextual cues~\cite{lee2019context,kosti2019context,le2022global,zhao2021robust}. Most approaches adopt a two-stream architecture to extract face and context features before fusing them for emotion prediction. Despite improvements, these models suffer from \textit{context bias}, where spurious correlations between background context and emotion (e.g., associating ``hospital'' with ``sadness'') lead to misclassifications.

Recent works tackle this bias using causal frameworks such as CCIM~\cite{yang2024towards} and CLEF~\cite{yang2024robust}. 
CCIM build a computationally expensive confounder dictionary by clustering contexts (e.g., ``hospital'', ``beach'') to reduce context bias, but it applies uniform adjustments across each representative context, potentially suppressing fine-grained emotional cues within similar-looking but emotionally diverse contexts.
CLEF avoids the dictionary but performs context debiasing after face-context fusion, using two parallel pipelines for factual and counterfactual, limiting its ability i) to refine context representations in isolation and ii) model the complex interactions between facial and contextual cues.

To overcome these limitations, we propose the \textbf{Attention Guided Context Debiasing Network (AGCD-Net)}, which performs instance-level correction of context features before fusion. AGCD-Net introduces a \textit{Hybrid ConvNeXt} encoder with integrated Spatial Transformer Networks (STN) and Squeeze-and-Excitation (SE) blocks for robust and aligned feature extraction. It further incorporates a dual-stream attention mechanism and the novel \textbf{Attention Guided - Causal Intervention Module (AG-CIM)}, which isolates and suppresses spurious context features under guidance from face embeddings. Unlike CCIM~\cite{yang2024towards}, AG-CIM performs targeted refinements instead of uniform corrections; unlike CLEF~\cite{yang2024robust}, it adjusts context representations before fusion, avoiding post-hoc bias suppression. The key contributions of our work are as follows:
\begin{enumerate}
    \item \textbf{Attention-based Dual Encoding Mechanism}: Enhances recognition accuracy by independently encoding face and context features using a robust Hybrid ConvNeXt architecture.
    \item \textbf{Causal Intervention Module}: Dynamically adapts and debiases context features based on facial information, reducing spurious correlations.
    \item \textbf{End-to-End Framework}: Integrates encoding, causal intervention, and feature fusion for seamless optimization.
    \item \textbf{State-of-the-Art Performance}: Demonstrates superior accuracy of the proposed AGCD-Net compared to existing methods.
\end{enumerate}

\section{Related Work}
\label{sec:soa}

Facial Emotion Recognition (FER) has evolved from early methods using handcrafted features like FACS~\cite{wang2013capturing} and LBPs~\cite{shan2009facial} to more advanced deep learning approaches such as CNNs, RNNs, and 3D-CNNs for capturing spatial and temporal information in static FER~\cite{fabian2016emotionet} and dynamic FER~\cite{jiang2020dfew,kossaifi2020factorized}. 
While these traditional techniques were effective in controlled settings, they struggled with real-world challenges such as pose variations, lighting conditions, occlusions and limited contextual awareness (overlooking the broader context surrounding the subject)~\cite{el2023emonext,bohi2024novel}. 

To address these limitations, Context-Aware Emotion Recognition (CAER) emerged, integrating facial and contextual features for more robust emotion analysis in diverse settings. Early work by~\emph{Kosti et al.}~\cite{kosti2019context} introduced CAER using a two-stream Convolutional Neural Network (CNN) for encoding face and context features. Later models enhanced this framework using attention mechanisms\cite{zhang2019context,le2022global} or multimodal fusion with body cues and scene understanding~\cite{li2021sequential}. Despite these advancements, CAER models often face challenges with \emph{context bias}, where over-reliance on background context leads to misinterpretation of emotional states, underscoring the need for methods to address this issue. 

To tackle this problem, causal theory-based frameworks have been introduced. These approaches aim to reduce contextual bias by identifying and separating meaningful context features from confounding factors. The Contextual Causal Intervention Module (CCIM)~\cite{yang2024towards} utilizes a confounder dictionary to learn this separation. This dictionary is built using an unsupervised learning method to identify the clusters of context feature by masking out the face-specific attributes in images. 
However, this framework requires significant computational resources, increasing model complexity for the generation of the confounder dictionary. Moreover, this method is sub-optimal, as the predefined intervention treats all contexts uniformly, making it difficult to distinguish beneficial cues from harmful ones. 
While CLEF~\cite{yang2024robust} adopts a parallel structure involving factual (face+context) and counterfactual (context-only) branches.  The framework then compares the predictions from \textit{factual} branch with the counterfactual predictions to isolate the harmful bias and debias final emotion predictions. Such a late-stage intervention limits the model’s capacity to disentangle contextual bias at a granular level and makes it harder to remove spurious contextual influences without also affecting useful emotional cues.

In contrast to these methods, our proposed AGCD-Net addresses context bias through early face-guided correction of context features, before fusion, ensuring that misleading correlations are removed while preserving semantically rich cues. This dynamic and attention-driven strategy overcomes the limitations of rigid prototypes and post-hoc adjustments, offering a more flexible and interpretable solution for real-world emotion recognition.

\section{Preliminary}
\label{sec:causal}
To build a strong foundation for the proposed model, we begin by introducing essential concepts from~\emph{causal inference theory}, which aims to analyze the affect of change in one variable (light or context) on final output (classification) while keeping other variables (subject/face) constant.
To get this understanding, we look at two scenarios:
\begin{enumerate}
    \item \textbf{Factual}: This scenario represents what is actually observed (face) in a specific setting (context/background).
    \item \textbf{Counterfactual}: This represents “what if” scenario. The face placed in a different context to assess the effect of background changes. The aim is to generate counterfactual to eliminate spurious correlations in data~\cite{pearl1993bayesian}.
\end{enumerate}

Comparing these scenarios, reveals how much the context biases emotion recognition. A large shift suggests a \emph{bias} rooted in the context rather than the facial expression. To address this, we apply causal adjustment only to the biased variable (i.e. context features) while keeping facial features unchanged to isolate context influence. This principle ensures that we isolate the effect of context alone.

The core research problem we address using causal theory is:\emph{ How can we reduce context-induced bias in emotion recognition by adjusting the causal influence of biased context features prior to fusion of face and context features}. The objective of AG-CIM module is 
to perform fine-grained, instance-specific refinement of context representations in isolation (prior to fusion), while preserving informative contextual semantics, ensuring that only meaningful context contributes to the final prediction.

\section{Proposed Method}
\label{sec:method}
To tackle the issue of context bias in context aware emotion recognition, we proposed Attention Guided Context Debiasing Network (AGCD-Net), a model designed for robust emotion recognition in dynamic environment. As context and facial cues, both are important for FER, AGCD-Net extracts facial and contextual cues using an~\textbf{Attention-Based Dual Encoding Network} while actively correcting for biased context through an \textbf{Attention Guided - Causal Intervention Module (AG-CIM)} before fusing them for final classification.   

\begin{figure*}[h]
    \centering
    \includegraphics[width=1\linewidth]{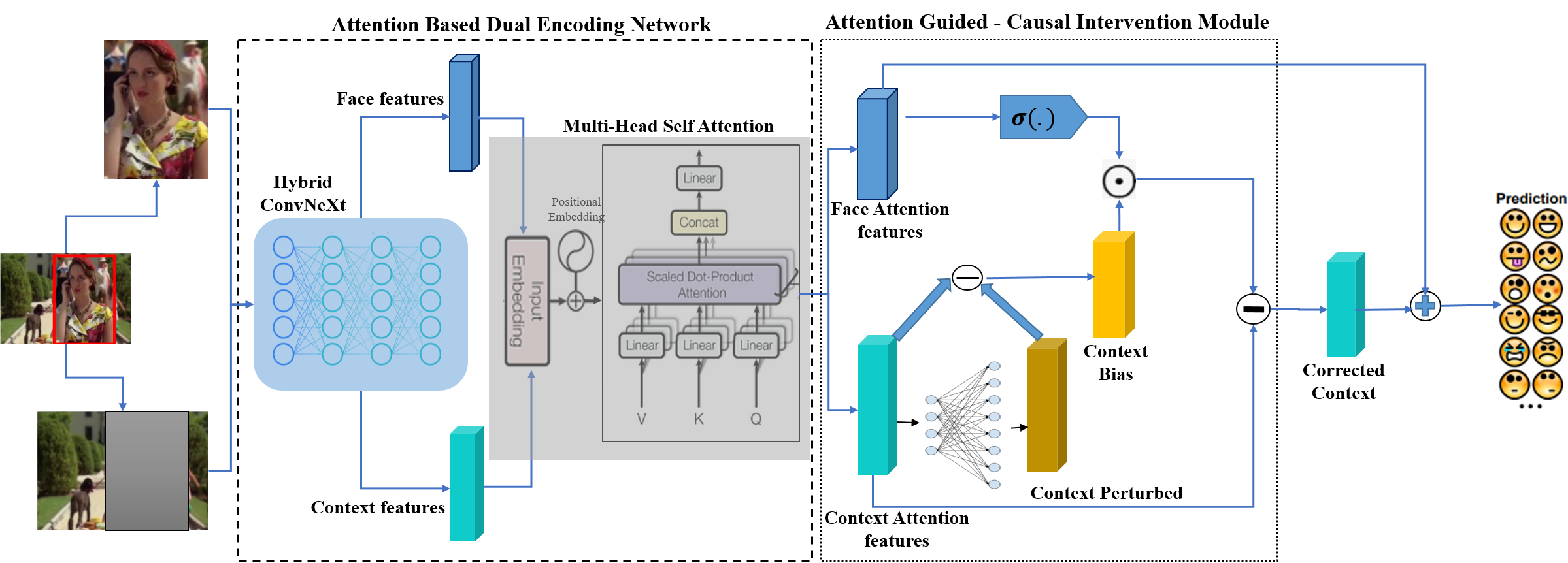}
    \caption{Proposed Attention Guided Context Debiasing Model for Context-Aware Emotion Recognition}
    \label{fig:methodology}
\end{figure*}
Specifically, let an image be represented as $ I $ with a discrete emotion label $ y $ from a set of $ K $ emotions $ \{ y_1, \ldots, y_K \} $. Our goal is to predict the emotion label $ y $ by processing separate facial and contextual features, modeled as $ \phi_f $ and $ \phi_c $, respectively. To achieve this, we propose AGCD-Net with three components: i) \emph{Attention-Based Dual Encoding Network} for independently encoding and focusing on face and context, ii) \emph{Attention Guided - Causal Intervention Module} to correct contextual biases and eliminate influences from irrelevant context features, and  iii) \emph{Final Classification Module} to produce the predicted emotion based on the debiased and fused features, as illustrated in Figure~\ref{fig:methodology}.

\subsection{Attention-Based Dual Encoding Network}

The Attention-Based Dual Encoding Network is meticulously designed to process the face and context streams independently, thereby enabling the model to effectively distinguish between emotional cues derived from facial expressions and those arising from the surrounding context.

\subsubsection{Hybrid ConvNeXt: Encoder CNN for Feature Extraction.}

To extract robust features from facial and contextual images, we introduce Hybrid ConvNeXt, an enhanced version of the ConvNeXt architecture tailored for computer vision classification, with specific design elements aimed at improving context-aware emotion recognition. The enhancements focus on improving feature extraction and robustness to spatial variations in both facial and contextual features, as illustrated in Figure \ref{fig:hybrid_convnext}. The Core Components of Hybrid ConvNeXt are:

\begin{figure*}[ht]
    \centering
    \includegraphics[width=0.9\linewidth]{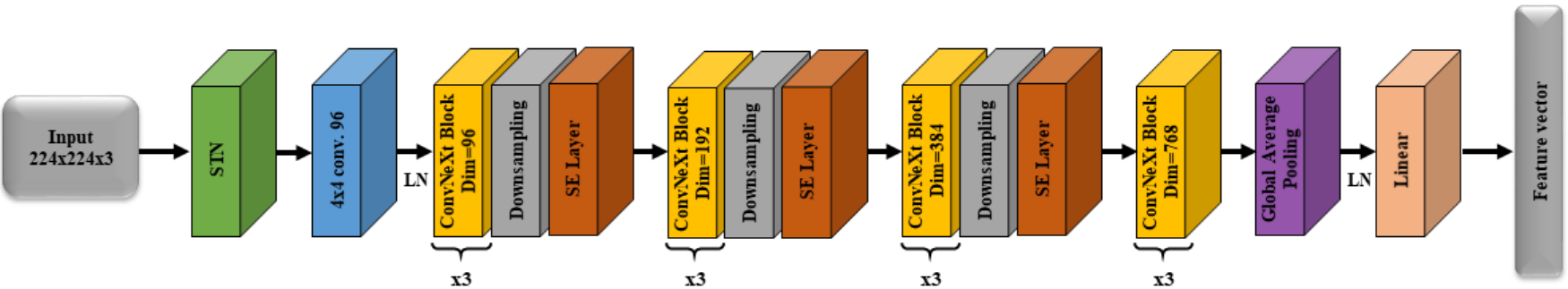}
    \caption{Proposed Hybrid ConvNeXt Encoding Network illustrating the key components such as ConvNeXt backbone, STN, and SE blocks}
    \label{fig:hybrid_convnext}
\end{figure*}

\begin{itemize}

    \item ConvNeXt Backbone: The foundation of Hybrid ConvNeXt is the hierarchical multi-stage design of ConvNeXt~\cite{liu2022convnet}, which employs patchified convolutions for downsampling, depthwise convolutions for efficient feature extraction, and advanced normalization techniques like LayerNorm. This structure enables the extraction of robust, multiscale spatial features. As illustrated in Figure~\ref{fig:convnext_block}, ConvNeXt leverages larger kernel sizes and replaces the ReLU activation function with GELU for smoother gradient flow, contributing to the improved performance of Hybrid ConvNeXt.

\begin{figure}
    \centering
    \subfigure[\label{fig:convnext_block}]{\includegraphics[width=0.17\linewidth]{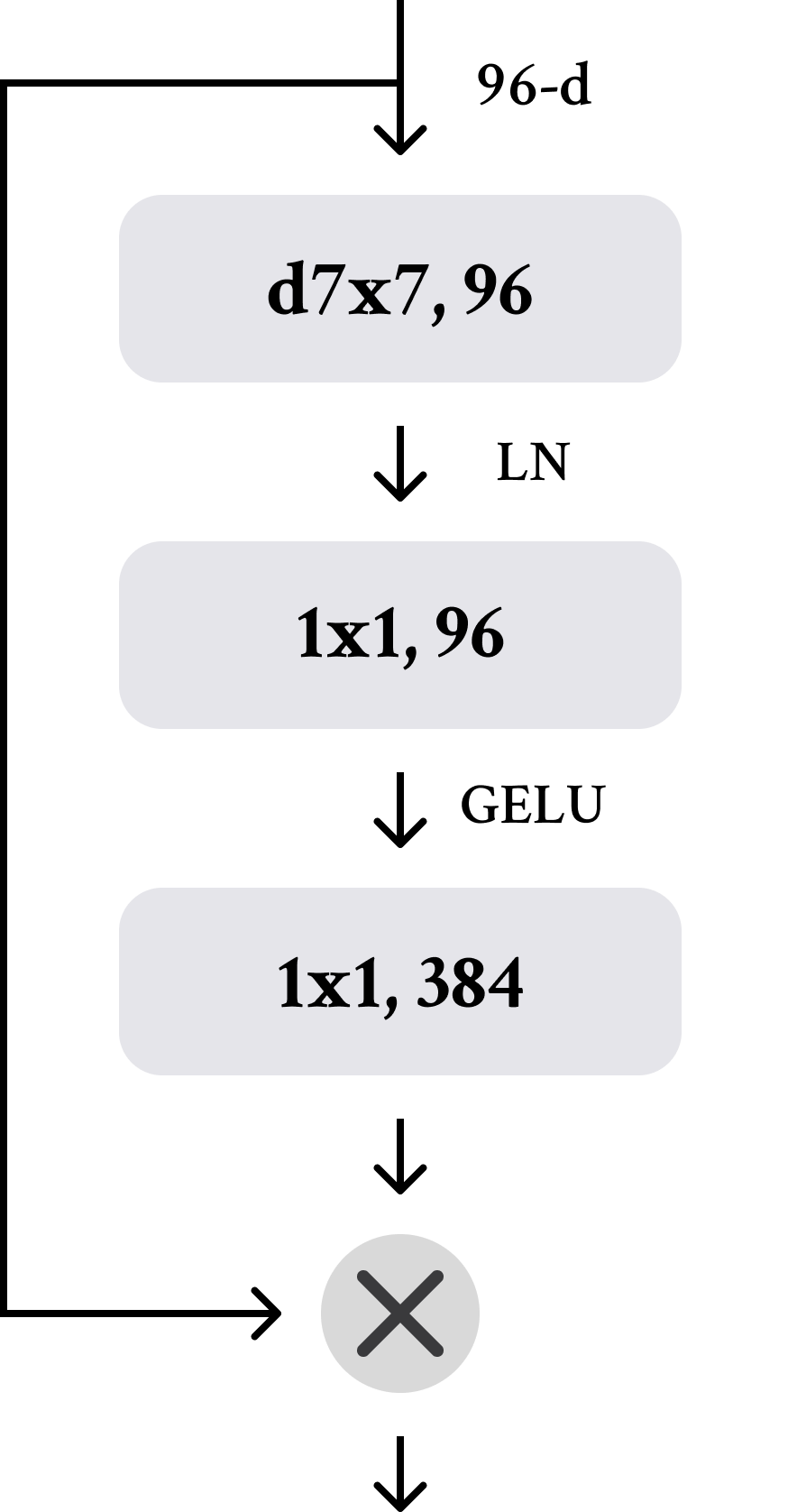}}
    \subfigure[\label{fig:stn_fig}]{
        \includegraphics[width=0.4\linewidth]{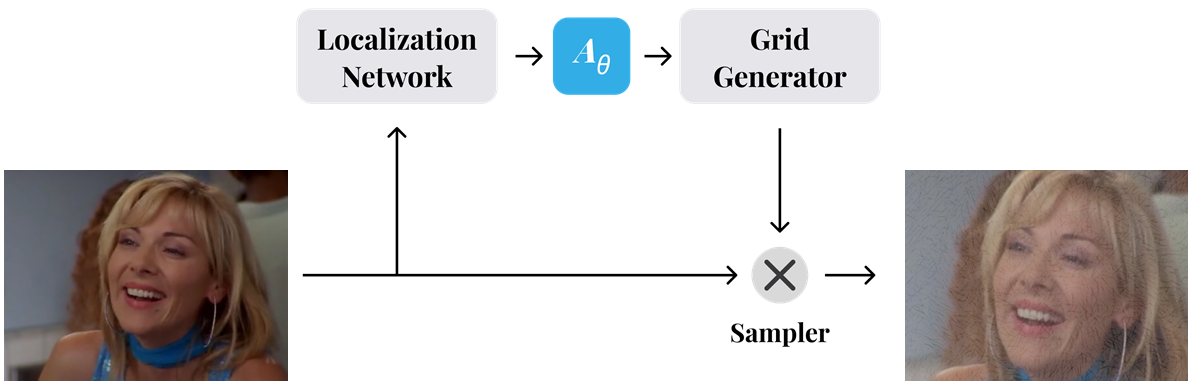}
    }
    \subfigure[\label{fig:se_block}]{\includegraphics[width=0.38\linewidth]{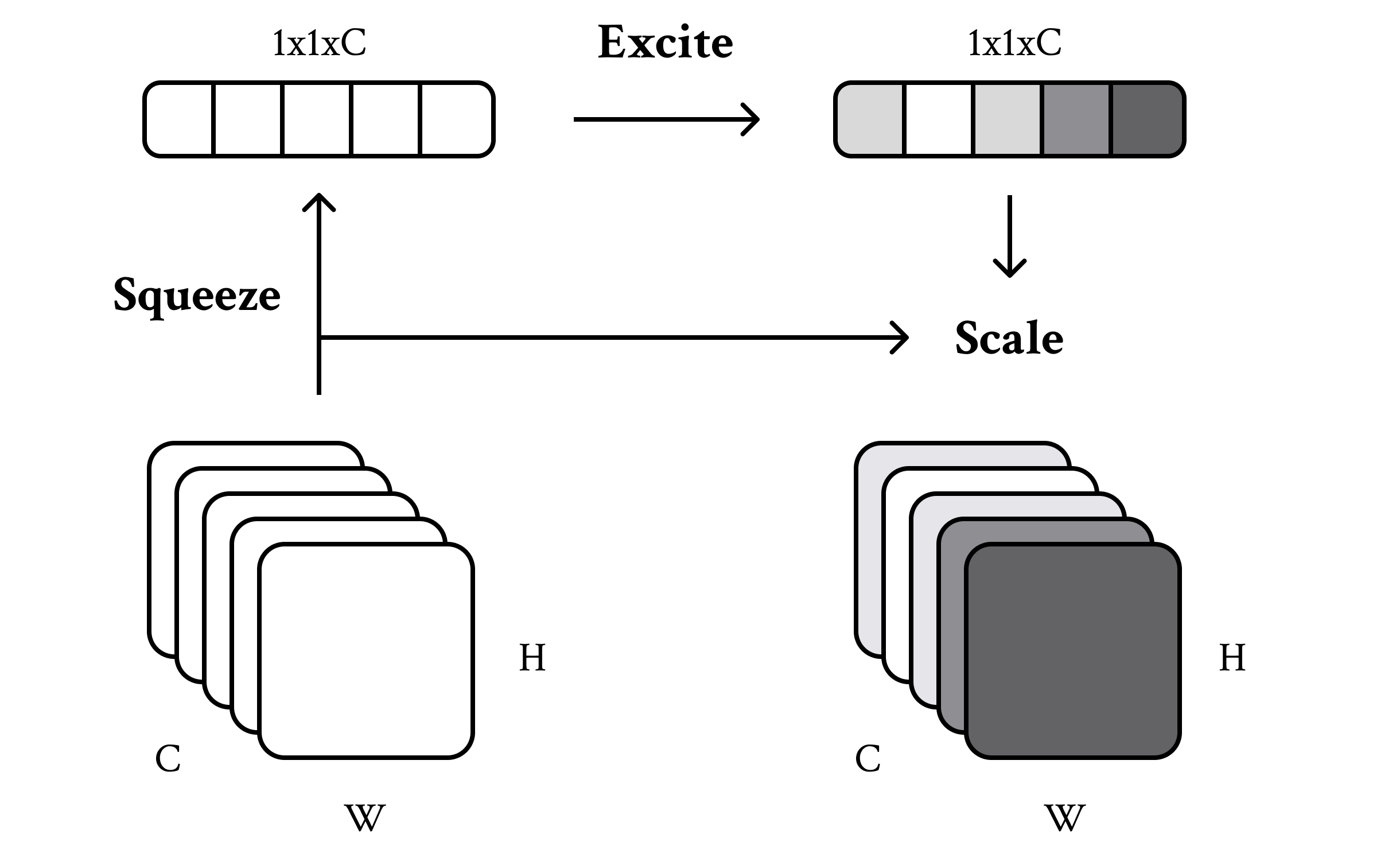}}
    \caption{Architectural components used in the Hybrid ConvNeXt: (a) ConvNeXt block, (b) Spatial Transformer module, and (c) Squeeze-and-Excitation block.}
    \label{fig:architecture_all}
\end{figure}

    \item Spatial Transformer Network (STN): An STN~\cite{jaderberg2015spatial} is integrated at the input to enhance the model's ability to handle spatial variations (scale, rotation, and translation) by learning spatial transformations directly from the data. STN consists of three key components, as illustrated in Figure~\ref{fig:stn_fig}:
    a \textbf{localization network} that predicts the transformation parameters, a \textbf{grid generator} that produces sampling coordinates, and a \textbf{sampler} that applies the transformation to the input image.
    The transformation parameters can be represented using a 3x2 affine transformation matrix $A_\theta$:
    \begin{equation}
        A_\theta =
        \begin{bmatrix}
            s_x \cos(\theta) & -s_y \sin(\theta) & t_x \\
            s_x \sin(\theta) & s_y \cos(\theta) & t_y 
        \end{bmatrix}
    \end{equation}
    where $t_x$, $t_y$ are the translation parameters, $s_x$, $s_y$ are the scale factors, and $\theta$ is the rotation angle. 
    The grid generator uses $A_\theta$ to produce a set of sampling grid coordinates, and the sampler applies these to the input image via bilinear interpolation, ensuring precise alignment of facial regions before feature extraction.

    \item Squeeze-and-Excitation (SE) Blocks: SE blocks~\cite{hu2018squeeze} are applied after the ConvNeXt feature extraction stages to recalibrate channel-wise feature responses adaptively. The SE mechanism comprises two main operations: squeeze and excitation, as illustrated in Figure~\ref{fig:se_block}.

    In the squeeze operation, global average pooling is applied to each channel to capture the global distribution:
    \begin{equation}
        z_c = \frac{1}{H \times W} \sum_{i=1}^H \sum_{j=1}^W X_{i,j,c}
    \end{equation}
    where $X_{i,j,c}$ represents the value of the $c$-th channel at spatial position $(i, j)$. 

    The excitation operation models the interdependencies among channels and recalibrates them using two fully connected layers with non-linear activations:
    \begin{equation}
        s = \sigma(W_2 \cdot \text{ReLU}(W_1 \cdot z))
    \end{equation}
    where $W_1$ and $W_2$ are weight matrices, $z$ is the descriptor vector from the squeeze operation, and $\sigma$ denotes the sigmoid function. The resulting scaling factors are applied to recalibrate the feature maps.

\end{itemize}
\subsubsection{Feature Extraction Process.}
For our context-aware emotion recognition system, Hybrid ConvNeXt serves as the feature encoder for both facial and contextual images. 
This design allows the network to leverage spatially and contextually aware feature embeddings from both the face and the background of each image. We ensure that the extracted features from face $(\phi_{f})$ and context ($\phi_{c}$) remain generalizable and ready for downstream processing in the next modules: the Multi-Head Self-Attention (MHSA) and the Attention Guided - Causal Intervention Module.

\subsubsection{Multi-Head Self Attention.}
Multi-Head Self-Attention (MHSA) enhances the model's ability to encode diverse information by allowing it to focus on multiple parts of the face and context features simultaneously.
This is achieved by linearly transforming the features from face $(\phi_{f})$ and context ($\phi_{c}$) into multiple query, key, and value vectors, and combining the outputs to produce attention-refined face ($\phi_{f,\text{att}}$) and context ($\phi_{c,\text{att}}$) features.

\subsection{Attention Guided - Causal Intervention Module (AG-CIM)}
The  Attention Guided Context Debiasing Network (AGCD-Net) model incorporates causal theory (Section~\ref{sec:causal}) to address spurious correlations in context features that often misleads prediction model. The Attention Guided - Causal Intervention Module (AG-CIM) is designed to correct these biases by simulating counterfactual scenarios and adjusting context features, using face features to guide the corrections.

\subsubsection{Counterfactual Generation.}
The first step of AG-CIM involves generating counterfactual representations of the context to simulate how the context features appear if the spurious correlation between context and emotion class were minimized. The attended context features $ \phi_{c,\text{att}} $ are perturbed through learnable transformation to generate the counterfactual representation, as shown in Figure~\ref{fig:methodology}. Mathematically it is represented as:

\begin{equation}
    \phi_{c,\text{pert}} = \mathbf{W}_p \cdot \phi_{c,\text{att}}
\end{equation}

where $\mathbf{W}_p$ is a learnable weight matrix trained via backpropagation to generate counterfactual context features by introducing controlled perturbations that suppress spurious or irrelevant patterns.

\subsubsection{Context Bias Computation.}
Next, the bias in context features is quantified as the difference between original and perturbed (counterfactual) features:
\begin{equation}
    \Delta \phi_c = \phi_{c,\text{att}} - \phi_{c,\text{pert}}
    \label{eq:bias_compute}
\end{equation}
This difference represents the spurious parts of the context that should be removed to prevent bias. If the difference $\Delta \phi_c$ is significant, it suggests that the context itself plays a major role in biasing the output (emotion prediction). By computing the difference between the original and perturbed context features, AGCD-Net identifies spurious correlations in the context and applies a controlled correction before making predictions. 

\subsubsection{Bias Correction Using Face Attention.}
To apply correction and decide how much weight to assign to context bias, we apply another linear transformation $ \mathbf{W}_c $ and then adjust it with a sigmoid function. 
If the bias (equation~\ref{eq:bias_compute}) is large, the sigmoid function will allow more correction; if it’s small, it dials down the correction. Additionally, a learnable parameter $\alpha$ is used to dynamically adjust the influence of face features to control their relative importance in the correction process as follows:

\begin{equation}
    \phi_{c,\text{corr}} = \phi_{c,\text{att}} - \mathbf{W}_c \Delta \phi_c \cdot \sigma(\alpha \times \phi_{f,\text{att}})
\end{equation}

where $ \mathbf{W}_c $ is a learnable parameter matrix. The $\alpha$ is learnable parameter initialized to $1$, and the sigmoid function learns and adjusts the strength of the correction based on the face features informativeness. This approach minimizes the impact of irrelevant context, reducing bias.

\subsection{Fusion and Classification}
The face features and unbiased context features are fused via element-wise addition and passed through fully connected layer and softmax function to estimate the probability distribution over emotion classes:

\begin{equation}
    \hat{y} = \text{softmax}(\mathbf{W}_f \cdot ( \phi_{f,\text{att}} + \phi_{c,\text{corr}}))
\end{equation}

where $\mathbf{W}_f$ maps the fused feature space to the emotion label space.
\newline To optimize the model's performance, we used Cross-Entropy Loss $(\mathcal{L}_{\text{CE}})$ for classification. Additionally, we introduce the \emph{\textbf{Attention Loss ($ \mathcal{L}_{\text{att}} $)}} to regularize the attention weights. The attention loss encourages balanced attention between the face and context features, defined as:

\begin{equation}
    \mathcal{L}_{\text{att}} = \frac{1}{N} \sum_{i=1}^{N} \Big( \lvert h_{\text{face}} \rvert + \lvert h_{\text{context}} \rvert \Big)
\end{equation}

where $ h_{\text{face}} $ and $ h_{\text{context}} $ represent the attention weights. This regularization term is added to the Cross-Entropy Loss, resulting in the final loss function:

\begin{equation}
    \mathcal{L}_{\text{final}} = \mathcal{L}_{\text{CE}} + \mathcal{L}_{\text{att}}
\end{equation}

\section{Experiments \& Results}
\label{sec:exp_results}
\subsection{Dataset \& Evaluation Metrics}
We used the CAER-S dataset with 70k images from 79 indoor and outdoor TV shows, covering diverse scenes and labeled across seven emotion classes. It was split into 70\% training, 10\% validation, and 20\% testing. Standard classification accuracy was used, following the official PyTorch implementation\footnote{{https://github.com/ndkhanh360/CAER}}.

\subsection{Implementation Details}
AGCD-Net was implemented in PyTorch using face and context features extracted by the proposed Hybrid ConvNeXt trained on CAER-S. Training employed AdamW optimizer (lr = $1 \times 10^{-5}$, weight decay = $1 \times 10^{-4}$), a CosineAnnealingWarmRestarts scheduler ($T_0{=}128$, $T_{mult}{=}2$), and a batch size of 128. Data augmentation was applied, and the loss combined cross-entropy with label smoothing ($\epsilon{=}0.2$) and a self-attention loss to enforce embedding consistency.

\subsection{Results \& Discussion}
\label{sec:results}

Table~\ref{tab:soa_compare} presents a clear comparison of baseline, SOTA, and ablation models (CIM modules \& AGCD-Net) on CAER-S.
The results demonstrate: 
i) the \textbf{superior performance of our proposed AGCD model} (ablation - last row), achieving 90.65\% accuracy, and 
ii) the \textbf{effectiveness of our proposed Hybrid ConvNeXt model} \textit{without the context debiasing (AG-CIM) module} (SOTA - last row), achieving 88.84\%, compared to existing state-of-the-art CAER models. 
Additionally, combining CAER-Net~\cite{lee2019context} with our AG-CIM (Ablation - $2^{nd}$ last row), achieved a notable accuracy of 76.03\%, representing improvements of $1.63\%$ and $0.22\%$ compared to CCIM~\cite{yang2024towards} and CLEF~\cite{yang2024robust}, respectively.
This shows the effectiveness of Hybrid ConvNeXt and face-guided, early-stage context debiasing over rigid and post-hoc prototypes like CCIM~\cite{yang2024towards} and CLEF~\cite{yang2024robust}.

\begin{table*}[!ht]
\centering
\caption{Performance comparison of CAER models on the CAER-S test set, including pre-trained ImageNet models on CAER-S, fine-tuned ImageNet models on CAER-S, SOTA, and ablation results with Causal Intervention Modules.}

\resizebox{0.8\linewidth}{!}{
\begin{tabular}{cccc}
\hline
Category & Methods                                   & Year & Accuracy (\%)  \\ \hline \hline
\multirow{3}{*}{Pre-trained} & ImageNet-AlexNet~\cite{krizhevsky2012imagenet}   & 2012 & 47.36 \\
         & ImageNet-VGGNet~\cite{simonyan2014very}   & 2015 & 49.89          \\
         & ImageNet-ResNet~\cite{he2016deep}         & 2016 & 57.33          \\ \hline
\multirow{3}{*}{Fine-tuned}  & Fine-tuned AlexNet~\cite{krizhevsky2012imagenet} & 2012 & 61.73 \\
         & Fine-tuned VGGNet~\cite{simonyan2014very} & 2015 & 64.85          \\
         & Fine-tuned ResNet~\cite{he2016deep}       & 2016 & 68.46          \\ \hline
\multirow{7}{*}{SOTA}        & CAER-Net-S~\cite{lee2019context}                 & 2019 & 73.52 \\
         & EMOT-Net~\cite{kosti2019context}          & 2019 & 74.51          \\
         & GNN-CNN~\cite{zhang2019context}           & 2019 & 77.21          \\
         & GLAMOR-Net~\cite{le2022global}          & 2022 & 77.90          \\
         & CAHFW-Net~\cite{zhou2023emotion}          & 2023 & 83.76          \\
         & EfficientFace~\cite{zhao2021robust}       & 2021 & 85.87          \\
         & \textbf{Hybrid ConvNeXt w/o AG-CIM (Ours)}         & 2025 & \textbf{88.84} \\ \hline
\multirow{4}{*}{Ablation}    & CAER-Net-S + CCIM~\cite{yang2024towards}           & 2024 & 74.81 \\
         & CAER-Net-S + CLEF~ \cite{yang2024robust}    & 2024 & 75.86          \\
         & \textbf{CAER-Net-S + AG-CIM (Our)}                   & 2025 & \textbf{76.03} \\
         & \textbf{AGCD (ours)}                               & 2025 & \textbf{90.65} \\ \cline{1-4} 
\end{tabular}}

\label{tab:soa_compare}
\end{table*}

\begin{figure}[!ht]
    \centering
    \includegraphics[width=0.55\linewidth]{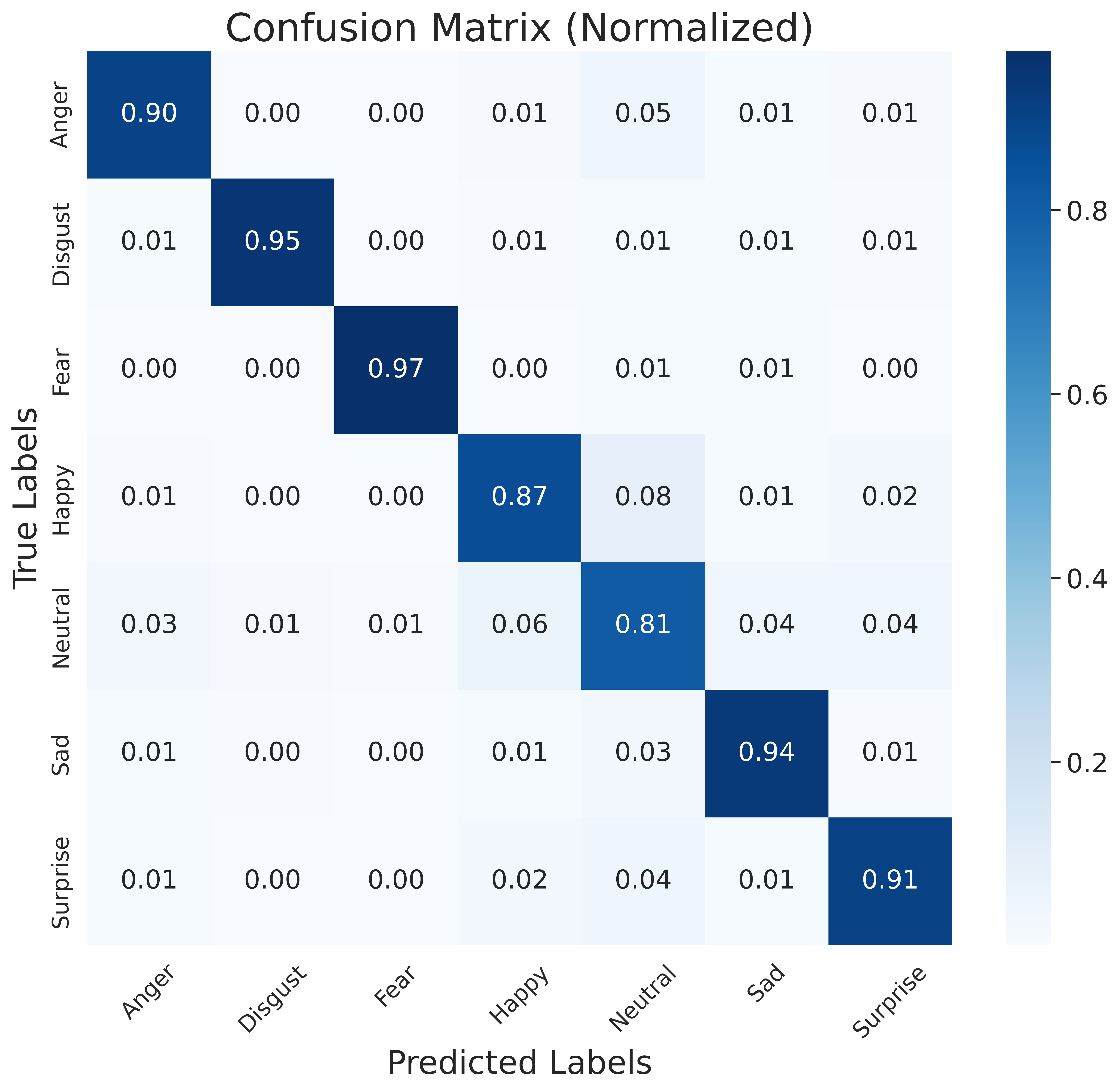}
    \caption{Normalized Confusion Matrix of AGCD-Net on CAER-S Dataset}
    \label{fig:conf_matrix}
\end{figure}

The normalized confusion matrix, shown in Figure~\ref{fig:conf_matrix}, evaluates the model's performance across emotion categories. The diagonal values represent high classification accuracy, with Fear (97\%) and Disgust (95\%) achieving the best results. 
However, the model struggles to predict the Neutral (81\%) and Happy (87\%) classes due to significant misclassification between the two. This issue arises from the high correlation (facial and contextual similarities) between these emotions in the dataset, leading to overlapping feature spaces. The highest off-diagonal values in the Neutral column correspond to Happy and vice versa, underscoring the challenge. Addressing this issue, possibly through more refined data collection for these classes in addition to the data augmentation techniques already implemented in our model, could significantly enhance prediction accuracy.

\subsection{Ablation Study}
\begin{table*}[!ht]
\centering
\caption{Ablation study on the impact of AGCD-Net components. $\checkmark$ = active, $\times$ = inactive.}

\resizebox{0.7\linewidth}{!}{ 
\begin{tabular}{ccccc}
\hline
Model & Face MHS- & Context MHS- & Causal Intervention & Overall Accuracy \\
      & Attention & Attention    & Module (AG-CIM)             & (in \%)          \\ \hline \hline
A & \checkmark & \checkmark & \checkmark & 90.65 \\ 
B & $\times$   & $\times$   & \checkmark & 90.40 \\
C & \checkmark & \checkmark & $\times$   & 90.27 \\
D & $\times$   & \checkmark & \checkmark & 90.47 \\ 
E & $\times$   & $\times$   & $\times$   & 88.84 \\ \hline
\end{tabular}}
\label{tab:ablation}
\end{table*}
Table~\ref{tab:ablation} evaluates the impact of key components on model performance by systematically removing or including the Face MHS-Attention, Context MHS-Attention, and the AG-CIM. \textbf{Model A}, including all components, achieves the highest accuracy of 90.65\%. \textbf{Model B}, without MHS-Attention but with AG-CIM, slightly decreases to 90.40\%, showing the robustness of AG-CIM. \textbf{Model C}, with both MHS-Attentions but without AG-CIM, reaches 90.27\%, indicating AG-CIM’s beneficial boost. \textbf{Model D}, missing Face MHS-Attention but retaining Context MHS-Attention and AG-CIM, performs at 90.47\%, highlighting i) the effectiveness of context attention and causal intervention, and ii) the strength of Hybrid ConvNeXt. \textbf{Model E}, using only Hybrid ConvNeXt without MHS-Attention or AG-CIM, records 88.84\%, still surpassing state-of-the-art techniques (Table~\ref{tab:soa_compare}), emphasizing the importance of these components.

\section{Conclusion and Future Directions}
\label{sec:conclusion}
We introduced AGCD-Net, an Attention Guided Context Debiasing model designed for emotion recognition in uncontrolled and dynamic environments. Leveraging the Hybrid ConvNeXt model with a dual attention mechanism, our framework extracts high-level context-aware features that outperform existing state-of-the-art methods. By integrating the proposed Attention-Guided Causal Intervention Module (AG-CIM), AGCD-Net effectively reduces context-induced bias and enhances classification accuracy. Extensive experiments on the CAER-S dataset demonstrate that AGCD-Net achieves state-of-the-art performance, significantly improving emotion recognition in challenging scenarios. These results underscore the effectiveness of causal debiasing and attention-driven correction in addressing spurious correlations. As future work, we plan to validate AGCD-Net on additional benchmarks, explore lightweight variants for edge deployment, and adapt it to healthcare scenarios involving emotionally vulnerable or minimally communicative individuals—such as patients with cognitive impairments.

\bibliographystyle{splncs04}
\bibliography{main}

\begin{thebibliography}{10}
\providecommand{\url}[1]{\texttt{#1}}
\providecommand{\urlprefix}{URL }
\providecommand{\doi}[1]{https://doi.org/#1}

\bibitem{ahmed2019emotion}
Ahmed, F., Bari, A.H., Gavrilova, M.L.: Emotion recognition from body movement. IEEE Access  \textbf{8},  11761--11781 (2019)

\bibitem{bohi2024novel}
Bohi, A., Boudouri, Y.E., Sfeir, I.: A novel deep learning approach for facial emotion recognition: application to detecting emotional responses in elderly individuals with alzheimer’s disease. Neural Computing and Applications pp. 1--19 (2024)

\bibitem{drimalla2020towards}
Drimalla, H., Scheffer, T., Landwehr, N., Baskow, I., Roepke, S., Behnia, B., Dziobek, I.: Towards the automatic detection of social biomarkers in autism spectrum disorder: Introducing the simulated interaction task (sit). NPJ digital medicine  \textbf{3}(1), ~25 (2020)

\bibitem{el2023emonext}
El~Boudouri, Y., Bohi, A.: Emonext: an adapted convnext for facial emotion recognition. In: 2023 IEEE 25th International Workshop on Multimedia Signal Processing (MMSP). pp.~1--6. IEEE (2023)

\bibitem{fabian2016emotionet}
Fabian Benitez-Quiroz, C., Srinivasan, R., Martinez, A.M.: Emotionet: An accurate, real-time algorithm for the automatic annotation of a million facial expressions in the wild. In: Proceedings of the IEEE conference on computer vision and pattern recognition. pp. 5562--5570 (2016)

\bibitem{greco2019emotion}
Greco, A., Roberto, A., Saggese, A., Vento, M., Vigilante, V.: Emotion analysis from faces for social robotics. In: 2019 IEEE international conference on systems, man and cybernetics (SMC). pp. 358--364. IEEE (2019)

\bibitem{gursesli2024facial}
Gursesli, M.C., Lombardi, S., Duradoni, M., Bocchi, L., Guazzini, A., Lanata, A.: Facial emotion recognition (fer) through custom lightweight cnn model: performance evaluation in public datasets. IEEE Access  (2024)

\bibitem{he2016deep}
He, K., Zhang, X., Ren, S., Sun, J.: Deep residual learning for image recognition. In: Proceedings of the IEEE conference on computer vision and pattern recognition. pp. 770--778 (2016)

\bibitem{hu2018squeeze}
Hu, J., Shen, L., Sun, G.: Squeeze-and-excitation networks. In: Proceedings of the IEEE conference on computer vision and pattern recognition. pp. 7132--7141 (2018)

\bibitem{jaderberg2015spatial}
Jaderberg, M., Simonyan, K., Zisserman, A., et~al.: Spatial transformer networks. Advances in neural information processing systems  \textbf{28} (2015)

\bibitem{jiang2020dfew}
Jiang, X., Zong, Y., Zheng, W., Tang, C., Xia, W., Lu, C., Liu, J.: Dfew: A large-scale database for recognizing dynamic facial expressions in the wild. In: Proceedings of the 28th ACM international conference on multimedia. pp. 2881--2889 (2020)

\bibitem{kossaifi2020factorized}
Kossaifi, J., Toisoul, A., Bulat, A., Panagakis, Y., Hospedales, T.M., Pantic, M.: Factorized higher-order cnns with an application to spatio-temporal emotion estimation. In: Proceedings of the IEEE/CVF conference on computer vision and pattern recognition. pp. 6060--6069 (2020)

\bibitem{kosti2019context}
Kosti, R., Alvarez, J.M., Recasens, A., Lapedriza, A.: Context based emotion recognition using emotic dataset. IEEE transactions on pattern analysis and machine intelligence  \textbf{42}(11),  2755--2766 (2019)

\bibitem{krizhevsky2012imagenet}
Krizhevsky, A., Sutskever, I., Hinton, G.E.: Imagenet classification with deep convolutional neural networks. Advances in neural information processing systems  \textbf{25} (2012)

\bibitem{le2022global}
Le, N., Nguyen, K., Nguyen, A., Le, B.: Global-local attention for emotion recognition. Neural Computing and Applications  \textbf{34}(24),  21625--21639 (2022)

\bibitem{lee2019context}
Lee, J., Kim, S., Kim, S., Park, J., Sohn, K.: Context-aware emotion recognition networks. In: Proceedings of the IEEE/CVF international conference on computer vision. pp. 10143--10152 (2019)

\bibitem{li2021sequential}
Li, X., Peng, X., Ding, C.: Sequential interactive biased network for context-aware emotion recognition. In: 2021 IEEE International Joint Conference on Biometrics (IJCB). pp.~1--6. IEEE (2021)

\bibitem{liu2022convnet}
Liu, Z., Mao, H., Wu, C.Y., Feichtenhofer, C., Darrell, T., Xie, S.: A convnet for the 2020s. In: Proceedings of the IEEE/CVF conference on computer vision and pattern recognition. pp. 11976--11986 (2022)

\bibitem{pearl1993bayesian}
Pearl, J.: [bayesian analysis in expert systems]: comment: graphical models, causality and intervention. Statistical Science  \textbf{8}(3),  266--269 (1993)

\bibitem{schuller2009acoustic}
Schuller, B., Vlasenko, B., Eyben, F., Rigoll, G., Wendemuth, A.: Acoustic emotion recognition: A benchmark comparison of performances. In: 2009 IEEE workshop on automatic speech recognition \& understanding. pp. 552--557. IEEE (2009)

\bibitem{shan2009facial}
Shan, C., Gong, S., McOwan, P.W.: Facial expression recognition based on local binary patterns: A comprehensive study. Image and vision Computing  \textbf{27}(6),  803--816 (2009)

\bibitem{simonyan2014very}
Simonyan, K., Zisserman, A.: Very deep convolutional networks for large-scale image recognition. arXiv preprint arXiv:1409.1556  (2014)

\bibitem{wang2013capturing}
Wang, Z., Li, Y., Wang, S., Ji, Q.: Capturing global semantic relationships for facial action unit recognition. In: Proceedings of the IEEE International Conference on Computer Vision. pp. 3304--3311 (2013)

\bibitem{icaart25}
Yahyaoui, M.A., Oujabour, M., {Ben Letaifa}, L., Bohi, A.: Multi-face emotion detection for effective human-robot interaction. In: Proceedings of the 17th International Conference on Agents and Artificial Intelligence - Volume 1: ICAART. pp. 91--99. INSTICC, SciTePress (2025). \doi{10.5220/0013170300003890}

\bibitem{yan2024multimodal}
Yan, J., Li, P., Du, C., Zhu, K., Zhou, X., Liu, Y., Wei, J.: Multimodal emotion recognition based on facial expressions, speech, and body gestures. Electronics (2079-9292)  \textbf{13}(18) (2024)

\bibitem{yang2024towards}
Yang, D., Yang, K., Kuang, H., Chen, Z., Wang, Y., Zhang, L.: Towards context-aware emotion recognition debiasing from a causal demystification perspective via de-confounded training. IEEE Transactions on Pattern Analysis and Machine Intelligence  (2024)

\bibitem{yang2024robust}
Yang, D., Yang, K., Li, M., Wang, S., Wang, S., Zhang, L.: Robust emotion recognition in context debiasing. In: Proceedings of the IEEE/CVF Conference on Computer Vision and Pattern Recognition. pp. 12447--12457 (2024)

\bibitem{zhang2019context}
Zhang, M., Liang, Y., Ma, H.: Context-aware affective graph reasoning for emotion recognition. In: 2019 IEEE International Conference on Multimedia and Expo (ICME). pp. 151--156. IEEE (2019)

\bibitem{zhao2021robust}
Zhao, Z., Liu, Q., Zhou, F.: Robust lightweight facial expression recognition network with label distribution training. In: Proceedings of the AAAI conference on artificial intelligence. vol.~35, pp. 3510--3519 (2021)

\bibitem{zhou2023emotion}
Zhou, S., Wu, X., Jiang, F., Huang, Q., Huang, C.: Emotion recognition from large-scale video clips with cross-attention and hybrid feature weighting neural networks. International Journal of Environmental Research and Public Health  \textbf{20}(2), ~1400 (2023)

\end{thebibliography}
\end{document}